\documentclass[10pt, a4paper]{article}

\usepackage{lrec-coling2024} 

\usepackage[utf8]{inputenc} 
\usepackage[T1]{fontenc}    
\usepackage{hyperref}       
\usepackage{xurl}            
\usepackage{amsfonts}       
\usepackage{nicefrac}       
\usepackage{microtype}      

\hypersetup{
    breaklinks=true,   
    pdfusetitle=true  
}

\usepackage{booktabs}
\usepackage{threeparttable}
\usepackage{multirow}
\usepackage{multicol}
\usepackage{makecell}
\usepackage{tabularx}
\usepackage{enumitem}
\setlist{noitemsep}
\usepackage{xcolor}
\definecolor{cb-blue-green} {RGB}{ 0,  073,  073}
\definecolor{cb-burgundy}   {RGB}{146,   0,   0}

\newcommand{\lb}{{\color{cb-blue-green}{\{}}}
\newcommand{\rb}{{\color{cb-blue-green}{\}}}}

\usepackage{amssymb}
\usepackage{pifont}
\newcommand{\xmark}{\ding{55}}
\newcommand{\cmark}{\ding{51}}

\title{A Family of Pretrained Transformer Language Models for Russian}

\name{Dmitry Zmitrovich$^{1}$, Alexander Abramov$^{1}$, Andrey Kalmykov$^{1}$, \\  
\textbf{Maria Tikhonova$^{1}$, Ekaterina Taktasheva$^{2}$$^*$, Danil Astafurov$^{1}$,} 
\\ \textbf{Mark Baushenko$^{1}$, Artem Snegirev$^{1}$, Vitalii Kadulin$^{1}$, Sergey Markov$^{1}$},
\\ \textbf{Tatiana Shavrina$^{3}$$^*$\thanks{\ \ $^*$ Work done while at SaluteDevices.}, Vladislav Mikhailov$^{4*}$, and Alena Fenogenova$^{1}$}
}

\address{$^{1}$SaluteDevices, $^{2}$University of Edinburgh, $^{3}$Institute of Linguistics, RAS, $^{4}$University of Oslo \\
\textbf{Correspondence:} \href{mailto:alenush93@gmail.com}{alenush93@gmail.com}\\}

\abstract{
Transformer language models (LMs) are fundamental to NLP research methodologies and applications in various languages. However, developing such models specifically for the Russian language has received little attention. This paper introduces a collection of 13 Russian Transformer LMs, which spans encoder (ruBERT, ruRoBERTa, ruELECTRA), decoder (ruGPT-3), and encoder-decoder (ruT5, FRED-T5) architectures. We provide a report on the model architecture design and pretraining, and the results of evaluating their generalization abilities on Russian language understanding and generation datasets and benchmarks. By pretraining and releasing these specialized Transformer LMs, we aim to broaden the scope of the NLP research directions and enable the development of industrial solutions for the Russian language.
\\ \newline \Keywords{Russian language models, Russian language understanding, Russian language generation} }

\begin{document}

\maketitleabstract

\section{Introduction}
Transformer language models (LMs;~\citealp{vaswani2017attention}) have emerged as an essential component of state-of-the-art approaches for various natural language understanding and generation tasks. These LMs undergo pretraining in a self-supervised manner at scale on large text corpora before being adapted to a downstream task via finetuning, few-shot learning, and instruction tuning~\cite{ruder-etal-2019-transfer,bommasani2022opportunities,chowdhery2022palm,ouyang2022training,touvron2023llama}. Open access to the pretrained models' weights allows the community to accelerate research and develop efficient industrial solutions~\cite{wolf-etal-2020-transformers}. However, most of these LMs are developed for English, which imposes substantial constraints on the potential of the language technologies.

The community has addressed this problem by releasing massively multilingual LMs~\citep[e.g.,][]{conneau2019cross,conneau-etal-2020-unsupervised,liu-etal-2020-multilingual-denoising,xue-etal-2021-mt5,workshop2023bloom} and monolingual LMs for typologically diverse languages~\citep[e.g.,][]{polignano2019alberto,le-etal-2020-flaubert-unsupervised,delobelle-etal-2020-robbert,cui-etal-2020-revisiting,kutuzov-etal-2021-large}. Nowadays, there is still a lack of Transformer LMs developed specifically for the Russian Language.

This paper introduces a family of pretrained Transformers LMs for Russian, which spans a diverse set of model architectures. We offer Russian versions of the BERT~\cite{devlin-etal-2019-bert}, RoBERTa~\cite{liu2019roberta}, ELECTRA~\cite{clark2019electra}, GPT-3~\cite{brown2020language}, T5~\cite{raffel2020exploring}, and UL2~\cite{tay2022ul2} models in multiple sizes. We report the development of our LMs and focus on evaluating them on a suite of standard Russian language understanding and generation datasets and benchmarks. The results show that our LMs outperform their multilingual counterparts and related Russian Transformer LMs on most tasks, achieving state-of-the-art performance. The main \emph{contributions} are the following: 

\begin{enumerate}[leftmargin=1.65em,topsep=0.4em,itemsep=0em]
    \item  We pretrain and release 13 Transformer-based LMs for the Russian language: ruBERT-base\footnote{\href{https://huggingface.co/ai-forever/ruBert-base}{\texttt{hf.co/ai-forever/ruBERT-base}}}, ruBERT-large\footnote{\href{https://huggingface.co/ai-forever/ruBert-large}{\texttt{hf.co/ai-forever/ruBERT-large}}}, ruRoBERTa-large\footnote{\href{https://huggingface.co/ai-forever/ruRoberta-large}{\texttt{hf.co/ai-forever/ruRoBERTa-large}}}, ruELECTRA-small\footnote{\href{https://huggingface.co/ai-forever/ruElectra-small}{\texttt{hf.co/ai-forever/ruELECTRA-small}}}, ruELECTRA-medium\footnote{\href{https://huggingface.co/ai-forever/ruElectra-medium}{\texttt{hf.co/ai-forever/ruELECTRA-medium}}}, ruELECTRA-large\footnote{\href{https://huggingface.co/ai-forever/ruElectra-large}{\texttt{hf.co/ai-forever/ruELECTRA-large}}}, ruGPT-3-small\footnote{\href{https://huggingface.co/ai-forever/rugpt3small_based_on_gpt2}{\texttt{hf.co/ai-forever/ruGPT-3-small}}}, ruGPT-3-medium\footnote{\href{https://huggingface.co/ai-forever/rugpt3medium_based_on_gpt2}{\texttt{hf.co/ai-forever/ruGPT-3-medium}}}, ruGPT-3-large\footnote{\href{https://huggingface.co/ai-forever/rugpt3large_based_on_gpt2}{\texttt{hf.co/ai-forever/ruGPT-3-large}}}, ruT5-base\footnote{\href{https://huggingface.co/ai-forever/ruT5-base}{\texttt{hf.co/ai-forever/ruT5-base}}}, ruT5-large\footnote{\href{https://huggingface.co/ai-forever/ruT5-large}{\texttt{hf.co/ai-forever/ruT5-large}}}, FRED-T5-large\footnote{\href{https://huggingface.co/ai-forever/FRED-T5-large}{\texttt{hf.co/ai-forever/FRED-T5-large}}}, and FRED-T5-XL\footnote{\href{https://huggingface.co/ai-forever/FRED-T5-1.7B}{\texttt{hf.co/ai-forever/FRED-T5-XL}}}. The LMs have been released over the last few years under the MIT license.
    \item We conduct a series of experiments to evaluate the generalization abilities of our LMs on a wide range of tasks, including machine reading comprehension, natural language inference, word sense disambiguation, coreference resolution, acceptability classification, inappropriateness identification, text simplification, text summarization, and text detoxification. The evaluation codebase is publicly available~\footnote{\href{https://github.com/ai-forever/russian-lm-evaluation}{\texttt{github.com/ai\-forever/russian\-lm\-evaluation}}}.
\end{enumerate}


\section{Related Work}
\subsection{Multilingual Language Models}
Russian is well-represented in the pretraining corpus of various massively multilingual LMs, such as mBERT~\cite{devlin-etal-2019-bert}, XLM-R~\cite{conneau-etal-2020-unsupervised}, RemBERT~\cite{chung2021rethinking}, mBART~\cite{liu-etal-2020-multilingual-denoising}, mT5~\cite{xue-etal-2021-mt5}, XGLM~\cite{lin2022fewshot}, mGPT~\cite{shliazhko2022mgpt}, BLOOM~\cite{workshop2023bloom}, and mDeBERTa~\cite{he2023debertav3}, \emph{inter alia}. The multilingual LMs have significantly contributed to achieving notable results in standard NLP tasks for Russian and its related languages~\cite{arkhipov-etal-2019-tuning}. However, with the development of their monolingual counterparts (see \S\ref{subsec:russian-lms}), these LMs have primarily served as strong baselines for more complex Russian language understanding and generation tasks~\citeplanguageresource[e.g.,][]{shavrina-etal-2020-russiansuperglue,sakhovskiy2021rusimplesenteval,mikhailov-etal-2022-rucola}. 

\subsection{Russian Language Models}
\label{subsec:russian-lms}
DeepPavlov~\cite{burtsev-etal-2018-deeppavlov} pretrained one of the first monolingual BERT-based LMs for Russian. The model configurations include (i) the RuBERT-base model pretrained on the Russian Wikipedia and news corpora~\cite{kuratov2019adaptation}, (ii) the RuBERT-base-conversational model~\footnote{\href{https://huggingface.co/DeepPavlov/rubert-base-cased-conversational}{\texttt{hf.co/DeepPavlov/rubert\-base\-conversational}}} pretrained on OpenSubtitles~\citelanguageresource{lison-tiedemann-2016-opensubtitles2016} and social media texts, and (iii) a distilled version of RuBERT-base-conversational~\cite{kolesnikova2022knowledge}. Yandex released RuLeanALBERT\footnote{\href{https://huggingface.co/yandex/RuLeanALBERT}{\texttt{hf.co/yandex/RuLeanALBERT}}}, a Russian version of the ALBERT model~\cite{lan2020albert}, and YaLM-100B~\footnote{\href{https://huggingface.co/yandex/yalm-100b}{\texttt{hf.co/yandex/YaLM-100B}}}, the largest publicly available Russian LM. The LMs are pretrained on a corpus of web texts, Wikipedia articles, texts from the Taiga corpus~\citelanguageresource{shavrina2017methodology}, and other multiple sources.  

In line with these works, we have contributed to developing open-source Russian LMs of various model architectures, which are widely used within the Russian NLP community for research and development purposes~\citeplanguageresource[e.g.,][]{dementievarusse,artemova2022runne2022,shamardina2022findings}.

\section{Models}
\label{sec:models}
This section describes the model pretraining corpus,  architecture design, and pretraining details.

\begin{table}[t!]
\centering
    \resizebox{\columnwidth}{!}{
\begin{tabular}{lccccccc}
\toprule
\textbf{Model} & \textbf{Wikipedia (ru/en)} & \textbf{News} & \textbf{Books} & \textbf{C4} & \textbf{OpenSubtitles} & \textbf{Size}   \\
\midrule
ruBERT & \cmark/\xmark & \cmark & \xmark  & \xmark &  \xmark& 30GB \\
ruRoBERTa & \cmark/\xmark & \cmark& \cmark & \xmark  &  \xmark& 250GB \\
ruELECTRA  & \cmark/\xmark  & \cmark & \cmark& \xmark &  \cmark  &  70GB \\
ruGPT-3  & \cmark/\cmark & \cmark& \cmark&  \cmark &  \xmark& 450GB \\
ruT5 & \cmark/\xmark & \cmark& \cmark&  \cmark &  \xmark& 300GB \\
FRED-T5  & \cmark/\xmark & \cmark& \cmark&  \cmark &  \xmark& 300GB \\ \bottomrule
\end{tabular}}
\caption{The pretraining corpus statistics.}
\label{tab:pretraindata}
\end{table}

\subsection{Pretraining Corpus}
\label{sec:pretrain}
\paragraph{Data Collection}~\autoref{tab:pretraindata} summarizes the general statistics of our pretraining corpus. The corpus includes texts from various publicly available resources, which represent diverse domains:

\begin{itemize}[leftmargin=1.1em,topsep=0.4em,itemsep=0em]
    \item Wikipedia — a collection of general-domain texts from the Russian and English Wikipedia corpora. The Wikipedia articles are extracted from the corresponding dumps with the help of the WikiExtractor tool~\cite{Wikiextractor2015}.
    \item News — a collection of news articles from the Taiga corpus and the Lenta, Gazeta, and Interfax news sources from the corus\footnote{\href{https://github.com/natasha/corus/tree/master}{\texttt{github.com/natasha/corus}}} library.
    \item Books — a collection of literary texts from the librusec corpus~\citelanguageresource{Panchenko:17:aist} and poetic texts from the Taiga corpus. The texts are downloaded via the corus library.
    \item Colossal Clean Crawled Corpus (C4;~\citealp{raffel2020exploring}) — a collection of web texts in Russian. The C4 data is downloaded using the Tensorflow datasets~\cite{paper2021tensorflow}.
    \item OpenSubtitles — a collection of movie and TV subtitles extracted from parallel corpora.
\end{itemize}


\noindent In general, different domains and sizes of the sub-corpora are included in the resulting pretraining corpora of our LMs, which range from 30GB (ruBERT) to 450GB (ruGPT-3). This variability is primarily due to multiple factors. First, our models have undergone pretraining over a few years based on methodological advancements in developing LMs and creating pretraining corpora. For instance, the ruGPT-3's C4 sub-corpus differs from the ruT5 and FRED-T5 ones in that it is filtered according to the procedure described in~\citet{OrtizSuarezSagotRomary2019}. Second, the amount of textual data in the publicly available resources has increased over time, promoting an improved coverage of the world changes and domain representation. 

\begin{table*}[ht!]
\centering
    \scriptsize 
    \resizebox{\textwidth}{!}{\begin{tabular}{lccccccccc}
\toprule
\textbf{Model} & \textbf{Encoder} & \textbf{Decoder} & \textbf{Objective} & \textbf{Parameters} & \textbf{\# Layers} & $d_{model}$ & $d_{ff}$ & \textbf{Tokenizer} & \textbf{\# Heads} \\
\midrule
ruBERT-base & \cmark & \xmark & MLM \& NSP & 178M & 12 & 768 & 3072 & BPE, $12 \cdot 10^4$ & 12\\
ruBERT-large&  \cmark & \xmark & MLM \& NSP & 427M & 24 & 1024 & 4096 & BPE, $12 \cdot 10^4$ & 16\\ 
ruRoBERTa-large & \cmark & \xmark & MLM & 355M & 24 & 1024 & 4096 & BBPE, $5 \cdot 10^4$ & 16\\ 
ruELECTRA-small & \cmark & \xmark & RTD & 42M & 12 & 256 & 1024 & BPE, $256 \cdot 10^3$ & 4 \\
ruELECTRA-medium & \cmark & \xmark & RTD & 85M & 12 & 576 & 2304 & BPE, $64 \cdot 10^3$ & 12\\
ruELECTRA-large & \cmark & \xmark & RTD & 427M & 24 & 1024 & 4096 & BPE, $120 \cdot 10^3$ & 16\\ 
ruGPT-3-small & \xmark & \cmark  & LM & 125M & 12 & 768& 3072 & BBPE, $5 \cdot 10^4$  & 12\\
ruGPT-3-medium& \xmark & \cmark  & LM & 355M & 24 & 1024 & 4096 & BBPE, $5 \cdot 10^4$  & 16\\
ruGPT-3-large & \xmark & \cmark  & LM & 760M & 24 & 1536 & 6144 & BBPE, $5 \cdot 10^4$  & 16\\ 
ruT5-base & \cmark & \cmark & SP & 222M & 12 & 768& 3072 & SentencePiece, $32 \cdot 10^3$ & 12\\
ruT5-large & \cmark & \cmark &  SP & 737M & 24 & 1024 & 4096 & SentencePiece, $32 \cdot 10^3$ & 16\\ 
FRED-T5-large & \cmark & \cmark &  MoD & 820M & 24 & 1024 & 2816 &  BBPE, $5 \cdot 10^4$ & 16\\
FRED-T5-XL & \cmark & \cmark &  MoD & 1.74B & 24 & 1536 & 4096 & BBPE, $5 \cdot 10^4$ & 24 \\ \bottomrule
\end{tabular}}
\caption{Summary of the model architecture configurations. Pretraining objectives: language modeling (LM), masked language modeling (MLM), next sentence prediction (NSP), replaced token detection (RTD), span corruption (SP), and a mixture of denoisers (MoD). $d_{model}$ is the hidden layer dimension, and $d_{ff}$ is the feed-forward layer dimension. \textbf{Tokenizer} is the tokenization method and the vocabulary size.
}
\label{tab:architecture}
\end{table*}
\subsection{Architecture \& Pretraining Details}
The pretraining objectives, model architecture, scaling strategies, and other design choices for our LMs are summarized in~\autoref{tab:architecture}. The model configuration choices are based on extensive empirical studies described in detail in~\citet{devlin-etal-2019-bert,liu2019roberta,clark2020electra,brown2020language,tay2022ul2}, and other factors, such as availability 
of the data and computational resources, LM standards, and field state at a particular period of time, starting from the BERT model architecture.

\subsubsection{ruBERT}
\paragraph{Architecture} ruBERT is based on BERT~\cite{devlin-etal-2019-bert} and pretrained on (i) a masked language modeling (MLM) objective to predict masked-out tokens in the input and (ii) a next sentence prediction (NSP) objective to predict whether two sentences follow each other. We use two BERT versions (BERT-base and BERT-large) and the Byte-pair Encoding (BPE;~\citealp{wang2020neural}) tokenization, with the vocabulary size of $12 \cdot 10^4$ tokens. The main differences between DeepPavlov's ruBERT and our ruBERT LMs are the following. First, we pretrain and release the first ruBERT-large model. Second,  DeepPavlov's ruBERT models are pretrained with a small batch size on a limited number of GPUs. In contrast, we pretrain our ruBERT LMs on a similar pretraining corpus using a larger batch size and more computational resources, which results in improved model performance (see~\S\ref{sec:evaluation}).

 \paragraph{Pretraining Details} We pretrain ruBERT-base and ruBERT-large with a maximum sequence length of $512$ tokens using a linear scheduler with an initial learning rate of $1e{-4}$ and the Adam optimizer~\cite{kingma2017adam} with  $\beta_{1}=0.9$, $\beta_{2}=0.99$, and $\epsilon=1e^{-8}$. The masking probability is $0.15$. The total number of pretraining steps is $10^6$. ruBERT-base is pretrained for 
$8$ days on $16$ V100 GPUs, and ruBERT-large is pretrained for $20$ days on $16$ V100 GPUs.

\subsubsection{ruRoBERTa} 
\paragraph{Architecture} We use the RoBERTa-large configuration~\cite{liu2019roberta} for ruRoBERTa-large. The pretraining objective is MLM, the tokenization method is Byte-level BPE (BBPE;~\citealp{wang2020neural}), and the vocabulary counts $5 \cdot 10^4$ tokens.

\paragraph{Pretraining Details} We pretrain the model with a total batch size of $4096$, the maximum sequence length of $512$ tokens, a linear scheduler with an initial learning rate of $1e^{-4}$, and the Adam optimizer with $\beta_{1}=0.9$, $\beta_{2}=0.99$, and $\epsilon=1e^{-8}$. The masking probability is $0.15$. The model has seen 2T tokens during pretraining, which has taken 21 days on 64 V100 GPUs.

\subsubsection{ruELECTRA}
\paragraph{Architecture} We use the ELECTRA architecture configurations and follow the pretraining procedure described in~\citet{clark2020electra}. The models are pretrained with the replaced token detection (RTD) objective to predict which input tokens are masked by the MLM-based ``generator''. We use BPE with the vocabulary size of $256 \cdot 10^3$, $64 \cdot 10^3$, and $120 \cdot 10^3$ tokens for ruELECTRA-small, ruELECTRA-medium, and ruELECTRA-large, respectively.

 \paragraph{Pretraining Details} We pretrain the ruELECTRAmodels using the learning rate of $2e^{-4}$, the masking probability of $0.25$, the Adam optimizer with $\beta_{1}=0.9$, $\beta_{2}=0.99$, and $\epsilon=1e^{-6}$, and the maximum sequence length of $512$ tokens. ruELECTRA-small, ruELECTRA-medium, and ruELECTRA-large are pretrained with a batch size of $128$, $64$, and $48$ for $7$, $8$, and $10$ days on $4$ V100 GPUs for the total number of steps of $1 \cdot 10^6$, $1 \cdot 10^6$, and $4 \cdot 10^5$, respectively.

\subsubsection{ruGPT-3}
\paragraph{Architecture} ruGPT-3 is a Russian counterpart of GPT-3~\cite{brown2020language}. We use the model architecture description by~\citet{brown2020language} and the GPT-2 code base~\cite{radford2019language} from the Transformers library~\cite{wolf-etal-2020-transformers}. ruGPT-3 is pretrained on the language modeling objective. We use the BBPE tokenization with the vocabulary size of $5 \cdot 10^4$ tokens.

 \paragraph{Pretraining Details} The ruGPT-3 models are pretrained with a maximum sequence length of $1024$ tokens for three epochs and $2048$ tokens for one epoch. We use the initial learning rate of $1e^{-4}$ and the Adam optimizer with $\beta_{1}=0.9$, $\beta_{2}=0.99$, and $\epsilon=1e^{-8}$. The total number of tokens seen during pretraining is 80B. The pretraining of ruGPT3-small, ruGPT3-medium, and ruGPT3-large has taken 7, 16, and 16 days on 32, 64, and 128 V100-SXM3 GPUs, respectively.

\subsubsection{ruT5} 
\paragraph{Architecture} ruT5 is one of the first encoder-decoder LMs pretrained only on Russian-language textual data. ruT5 is designed analogically to T5~\cite{raffel2020exploring} and is available in two model configurations: ruT5-base and ruT5-large. The models are pretrained on an MLM span corruption objective, where consecutive spans of the input tokens are masked, and the model is trained to reconstruct the masked tokens. We use the SentencePiece tokenization~\cite{kudo-richardson-2018-sentencepiece} with the vocabulary size of  $32 \cdot 10^3$ tokens.

 \paragraph{Pretraining Details} The ruT5 models are pretrained using a linear scheduler with the learning rate of $1e^{-4}$ and the Adam optimizer with $\beta_{1}=0.9$, $\beta_{2}=0.99$, and $\epsilon=1e^{-8}$. The sequence length is set to 512/512 for inputs and targets. The ruT5-base and ruT5-large models are pretrained with a total batch size of $2048$ for 14 days on 32 V100 GPUs and 21 days on 64 V100 GPUs, respectively.

\subsubsection{FRED-T5} 
\paragraph{Architecture} FRED-T5 (Full-scale Russian Enhanced Denoisers) is an encoder-decoder model based on T5 and UL2~\cite{tay2022ul2}, available in two configurations: FRED-T5-large and FRED-T5-XL. In contrast to ruT5, FRED-T5 uses the gated GELU function instead of ReLU. Drawing inspiration from~\cite{tay2022ul2}, we pretrain FRED-T5 on a mixture of denoisers, a set of diverse pretraining objectives. The R-Denoiser is an MLM span corruption objective used in T5. The S-Denoiser follows the language modeling objective, where the input sequence is split into the context and target tokens so that the targets do not rely on future information. The X-Denoiser aims to recover much of the input based on the span corruption and language modeling objectives.

The main differences in the pretraining approaches between UL2 and FRED-T5 are the following: (i) we use seven denoisers with a uniform distribution of the hyperparameters $\mu$ (the average span length), $r$ (the corruption rate), and $n$ (the number of corrupted spans) instead of the normal distribution, and (ii) we use BBPE instead of SentencePiece, with a vocabulary size of $5 \cdot 10^4$ tokens.

We use the following special tokens and hyperparameters for the FRED-T5 denoisers: \texttt{<LM>} ($\mu=L/4$, $r=0.25$, $n=1$), \texttt{<SC1>} ($\mu=3$, $r=0.15$, $n=1$), \texttt{<SC2>} ($\mu=8$, $r=0.15$, $n=1$), \texttt{<SC3>} ($\mu=64$, $r=0.15$, $n=1$), \texttt{<SC4>} ($\mu=3$, $r=0.5$, $n=1$), \texttt{<SC5>} ($\mu=8$, $r=0.5$, $n=1$), \texttt{<SC6>} ($\mu=64$, $r=0.5$, $n=1$), where $L$ is the input length. The \texttt{<LM>} token corresponds to the S-Denoiser.

 \paragraph{Pretraining Details} FRED-T5 is pretrained using a linear scheduler with the initial learning rate of $1e^{-4}$ and the Adafactor optimizer~\cite{shazeer2018adafactor} with $\beta_{1}=0.9$, $\beta_{2}=0.99$, and $\epsilon=1e^{-8}$. The sequence length is set to 512/512 for inputs and targets. The FRED-T5-large and FRED-T5-XL models are pretrained with a total batch size of $2048$ for 35 days on 160 V100 GPUs, followed by 5 days on 80 A100 GPUs, and for 45 days on 112 A100 GPUs, respectively.

\section{Empirical Evaluation}
\label{sec:evaluation}
This section describes the experimental setup and presents the key results of evaluating our LMs on a suite of standard benchmarks and datasets for Russian. The optimal resulting hyperparameters are summarized in~\autoref{tab:hparams_finetune} (see \S\ref{app:hparams}).

\subsection{Natural Language Understanding}
\begin{table*}[h!]
    \centering
    \scriptsize 
    \resizebox{\textwidth}{!}{\begin{tabular}{@{}lccc@{ / }ccc@{ / }cccccc@{ / }c@{}}
    \toprule
    \multirow{2}{*}{\textbf{Model}} & \multirow{2}{*}{\textbf{Overall}} & \multicolumn{1}{c}{\textbf{LiDiRus}} & \multicolumn{2}{c}{\textbf{RCB}} &  \multicolumn{1}{c}{\textbf{PARus}} & \multicolumn{2}{c}{\textbf{MuSeRC}}  & \multicolumn{1}{c}{\textbf{TERRa}} & \multicolumn{1}{c}{\textbf{RUSSE}} & \multicolumn{1}{c}{\textbf{RWSD}} & \multicolumn{1}{c}{\textbf{DaNetQA}}  & \multicolumn{2}{c}{\textbf{RuCoS}} \\
    
    &   & \multicolumn{1}{c}{\textbf{MCC}} & \multicolumn{2}{c}{\textbf{F1/Acc.}} &  \multicolumn{1}{c}{\textbf{Acc.}} & \multicolumn{2}{c}{\textbf{F1$_a$/EM}} & \multicolumn{1}{c}{\textbf{Acc.}} & \multicolumn{1}{c}{\textbf{Acc.}} & \multicolumn{1}{c}{\textbf{Acc.}} & \multicolumn{1}{c}{\textbf{Acc.}} & \multicolumn{2}{c}{\textbf{F1/EM}} \\
    
    \midrule
    \multicolumn{14}{c}{\textbf{Encoder LMs}}\\
    \midrule

ruBERT-base      &60.3 &17.2   &35.7 &47.7 &70.4 &75.9 &41.4   &69.4 &73.9 &66.9&59.9   &85.0 &84.9   \\
ruBERT-large     &61.7 &20.1   &38.1 &49.3 &70.2 &79.4 &47.9   &70.5 &70.5 &66.9&67.8   &82.0 &82.0    \\
ruRoBERTa-large  &68.1 &34.1   &40.9 &46.3 &76.4 &84.5 &58.1   &79.3 &74.9 &66.9&81.1   &85.0 &85.0    \\
ruELECTRA-small  &50.5 &10.6   &34.6 &46.1 &56.4 &62.8 &21.0     &54.0 &59.2 &66.9&65.8   &60.0 &59.6   \\
ruELECTRA-medium &52.4 &  18.2  &41.3&52.5 &57.6 &61.5 &18.9   &54.4 &64.9 &66.9&60.0   &63.0 &62.4   \\
ruELECTRA-large  &52.2 &19.7  &38.6 &45.9 &64.4 &54.9 &7.8   &58.3 &63.2 &66.9&62.7   &61.0 &60.7  \\ 
ruBERT-base (DP)\textbf{*} & 57.6 & 19.9 & 26.5 & 45.7 & 54.2 & 77.7 & 43.3 & 64.8 & 71.4   &  66.9    & 60.1 &  84.0  & 84.0 \\
ruBERT-base-conv (DP)\textbf{*} & 50.0 & 17.8& 45.2 & 48.4 & 50.8 & 68.7 & 27.8 & 64.0 & 72.9 & 66.9  &  60.6 & 22.0  & 21.8 \\
mBERT\textbf{*} & 54.7 & 8.4 &  34.4 & 42.2 &  53.2 & 76.8 & 41.5 & 57.8 & 65.3 & 66.9 & 62.2 & 80.0 & 80.4 \\ 
XLM-R-large\textbf{*} & 63.9 & 35.1 & 32.3 & 46.8 & 51.0 & 81.5 & 50.7 & 79.1 & 77.0  & 66.9 & 73.7 & 86.0 & 86.3 \\  
RuLeanALBERT\textbf{*}   &    69.8 &     40.3 &  36.1 & 41.3 &   79.6 &  87.4 & 65.4 &   81.2 &   78.9 &  66.9 &     76.0 &  90.0 & 90.2 \\
FRED-T5-XL encoder-only\textbf{*}   &   69.4 &     42.1 &  31.1 & 44.1 &   80.6 &  88.2 & 66.6 &   83.1 &   72.3 &  66.9 &     73.5 &  91.0 & \underline{91.1} \\

    \midrule
    \multicolumn{14}{c}{\textbf{Decoder LMs}}\\
    \midrule

ruGPT-3-small & 43.8 & -1.3 & 35.6 & 47.3 & 56.2 & 65.3 & 22.1 & 48.8 & 57.0 & 66.9 & 61.0 &21.0 & 20.4 \\
ruGPT-3-medium & 46.8 & 1.0 & 37.2 & 46.1 & 59.8 & 70.6 & 30.8 & 50.5 & 64.2 & 66.9 & 63.4 &23.0 & 22.4 \\
ruGPT-3-large & 50.5 & 23.1 & 41.7 & 48.4 & 58.4 & 72.9 & 33.3 & 65.4 & 64.7 & 63.6 & 60.4 &21.0 & 20.2 \\
YaLM P-tune\textbf{*}  &    71.1 &     36.4 &  35.7 & 47.9 &   83.4 &  \underline{89.2} & \underline{70.7} &   84.1 &   71.0 &  66.9 &     85.0 &  \underline{92.0} & \textbf{91.6} \\
    \midrule
    \multicolumn{14}{c}{\textbf{Encoder-decoder LMs}}\\
    \midrule

ruT5-base     &62.3 &21.3 &42.5 &47.9 &57.8 &80.2 &47.1 &73.0  &71.3 &66.9 &76.9 &85.0 &84.8 \\
ruT5-large    &68.3 &35.1 &46.1 &51.6 &73.2 &84.9 &58.9 &77.9 &76.6 &66.9 &78.0  &86.0 &86.0  \\

FRED-T5-large &69.0  &33.8 &45.0 &48.4  &72.6 &88.0 &66.4  &79.6 &78.0  &66.9 &81.7 &85.0 &84.5 \\
FRED-T5-XL  &\underline{75.2} &\underline{46.5} &\underline{51.1} &\underline{54.6} &\underline{81.8} &\textbf{91.7} &\textbf{76.2} &\underline{86.9} &\textbf{81.7} &66.9 &\underline{88.2} &88.0 &88.0  \\

mT5-base\textbf{*} & 51.6 & 0.06 & 37.5 & 48.6 & 49.4 & 65.6 & 22.7 & 57.9 & 57.6 & 66.9 & 68.7 & 71.0 & 69.7 \\ 
mT5-large\textbf{*} & 56.0 & 17.0 & 34.4  &  42.7 & 50.4 & 77.6 & 42.9 & 67.3 & 56.4 & 66.9 &74.3& 74.0 & 72.8 \\ 

    \midrule
    Human  & \textbf{81.1} &  \textbf{62.6}&  \textbf{68.0}& \textbf{70.2} &   \textbf{98.2} &  80.6 & 42.0 &  \textbf{92.0}&   \underline{80.5} &  \textbf{84.0} &    \textbf{91.5} &  \textbf{93.0}& 89.0 \\
    \bottomrule
    \end{tabular}}
    \caption{Results on Russian SuperGLUE. All values are scaled by 100. DP=DeepPavlov~\cite{burtsev-etal-2018-deeppavlov}. \textbf{Overall} is the overall average score. The best score is in bold, and the second best is underlined. The baseline models are marked with an asterisk.}
    \label{tab:rsg_leaderboard}
    \end{table*}

\subsubsection{General Language Understanding}
\label{sec:rsg}
\paragraph{Tasks} Russian SuperGLUE~\citelanguageresource{shavrina-etal-2020-russiansuperglue} includes nine tasks on common sense understanding (RUSSE, PARus), natural language inference (TERRa, RCB), reasoning (RWSD), machine reading comprehension (MuSeRC, RuCoS;~\citealplanguageresource{fenogenova-etal-2020-read}) and world knowledge (DaNetQA;~\citealplanguageresource{glushkova2021danetqa}), and a broad-coverage diagnostic test set (LiDiRus). The \textbf{performance metrics} are the accuracy score (Acc.; PARus, TERRa, RUSSE, RWSD, RCB, and DaNetQA), exact match (EM; MuSeRC, RuCoS) the F1-score (F1; RCB, RuCoS), the macro-average F1-score (F1$_{a}$; MuSeRC), and the Matthews Correlation Coefficient (MCC; LiDiRus).

 \paragraph{Method} We estimate the model performance via finetuning and zero-shot evaluation. The encoder and encoder-decoder LMs are finetuned for a maximum of 40 epochs with an early stopping based on the task-specific performance metric or their average on the validation set. The task example templates are presented in~\autoref{tab:rsg_prompts} (see \S~\ref{app:rsg_templates}).

\begin{itemize}[leftmargin=1.1em,topsep=0.2em,itemsep=0em]
    \item Encoder LMs: we finetune the encoders 
    via the Transformers library using the AdamW optimizer~\cite{loshchilov2018decoupled}, learning rate of $1 \cdot 10^{-5}$, weight decay of $0.01$, and batch size of $32$.
    \item Decoder LMs: the decoder-only models are evaluated in a zero-shot setting, where the target label is selected based on the lowest perplexity of the resulting prompt templates. The ruGPT-3 results are taken from the official leaderboard as of September 2023: \href{https://russiansuperglue.com/leaderboard}{\texttt{russiansuperglue.com/leaderboard}}.
    \item Encoder-decoder LMs: we formulate the tasks in the text-to-text format and follow the two-stage finetuning procedure~\cite{raffel2020exploring}. The first stage is multi-task pretraining, where the model is continuously pretrained on a combination of tasks. Each input starts with a task-specific prefix. Next, the model is finetuned on each task individually using the bf16 precision. We experiment with using the combinations of Adam \& linear scheduler with a learning rate of $1 \cdot 10^{-5}$, and Adafactor \& constant scheduler with the learning rate of $1 \cdot 10^{-3}$.
\end{itemize}

 \paragraph{Baselines} We finetune ruBERT-base by DeepPavlov, mBERT, mT5-base, mT5-large and XLM-R-large as described above. We also 
compare our LMs with the following official leaderboard results: human annotators, ruBERT-base-conversational by DeepPavlov (ruBERT-base-conv), YaLM 3.3B \& P-tuning (YaLM P-tune), RuLeanALBERT, and the FRED-T5-XL encoder-only finetuned on each RSG task independently.

\begin{table*}[t!]
\scriptsize
\centering
\setlength{\tabcolsep}{12pt}
\begin{tabular}{@{}lcccccc@{}}
\toprule
\multirowcell{2}[-0.5ex][l]{\textbf{Model}} & \multicolumn{2}{c}{\textbf{Overall}} & \multicolumn{2}{c}{\textbf{In-domain}} & \multicolumn{2}{c}{\textbf{Out-of-domain}} \\ \cmidrule(lr){2-3} \cmidrule(lr){4-5} \cmidrule(lr){6-7}
 & \textbf{Acc.} & \textbf{MCC} & \textbf{Acc.} & \textbf{MCC} & \textbf{Acc.} & \textbf{MCC} \\ \midrule
\multicolumn{7}{c}{\textbf{Encoder LMs}} \\ \midrule

ruBERT-base & 74.50 \tiny{$\pm$ 0.60} & 0.41 \tiny{$\pm$ 0.01} & 76.95 \tiny{$\pm$ 0.72} & 0.36 \tiny{$\pm$ 0.01} & 73.17 \tiny{$\pm$ 0.74} & 0.43 \tiny{$\pm$ 0.01} \\

ruBERT-large & 75.90 \tiny{$\pm$ 0.42} & 0.42 \tiny{$\pm$ 0.01} & 78.82 \tiny{$\pm$ 0.57} & 0.40 \tiny{$\pm$ 0.01} & 74.30 \tiny{$\pm$ 0.71} & 0.42 \tiny{$\pm$ 0.01} \\

ruRoBERTa-large & \underline{80.80} \tiny{$\pm$ 0.47} & \underline{0.54} \tiny{$\pm$ 0.01} & \underline{83.48} \tiny{$\pm$ 0.45} & \underline{0.53} \tiny{$\pm$ 0.01} & \underline{79.34} \tiny{$\pm$ 0.57} & \underline{0.53} \tiny{$\pm$ 0.01} \\

ruELECTRA-small & 61.74 \tiny{$\pm$ 1.09} & 0.20 \tiny{$\pm$ 0.02} & 70.09 \tiny{$\pm$ 1.29} & 0.21 \tiny{$\pm$ 0.01} & 56.70 \tiny{$\pm$ 1.58} & 0.17 \tiny{$\pm$ 0.03} \\

ruELECTRA-medium & 74.11 \tiny{$\pm$ 0.85} & 0.38 \tiny{$\pm$ 0.02} & 76.14 \tiny{$\pm$ 0.88} & 0.34 \tiny{$\pm$ 0.02} & 73.00 \tiny{$\pm$ 1.05} & 0.38 \tiny{$\pm$ 0.02} \\

ruELECTRA-large & 65.65 \tiny{$\pm$ 0.65} & 0.20 \tiny{$\pm$ 0.02} & 72.79 \tiny{$\pm$ 0.31} & 0.22 \tiny{$\pm$ 0.01} & 61.75 \tiny{$\pm$ 1.02} & 0.17 \tiny{$\pm$ 0.02} \\

mBERT\textbf{*} & 67.47 \tiny{$\pm$ 1.33} & 0.19 \tiny{$\pm$ 0.01} & 72.69 \tiny{$\pm$ 1.40} & 0.19 \tiny{$\pm$ 0.02} & 64.63 \tiny{$\pm$ 1.62} & 0.18 \tiny{$\pm$ 0.02} \\ 

ruBERT-base (DP)\textbf{*} & 72.57 \tiny{$\pm$ 1.92} & 0.35 \tiny{$\pm$ 0.12} & 75.02 \tiny{$\pm$ 1.21} & 0.30 \tiny{$\pm$ 0.11} & 71.23 \tiny{$\pm$ 2.52} & 0.38 \tiny{$\pm$ 0.12} \\

ruBERT-base-conv (DP)\textbf{*} & 75.33 \tiny{$\pm$ 1.55} & 0.38 \tiny{$\pm$ 0.02} &  78.98 \tiny{$\pm$ 0.79} & 0.38 \tiny{$\pm$ 0.01} & 73.33 \tiny{$\pm$ 2.08} & 0.38 \tiny{$\pm$ 0.04} \\ 

RuLeanALBERT\textbf{*} & 80.00 \tiny{$\pm$ 0.0} & 0.52 \tiny{$\pm$ 0.0} & 82.00 \tiny{$\pm$ 0.0} & 0.49 \tiny{$\pm$ 0.0} & 78.00 \tiny{$\pm$ 0.0} & 0.52 \tiny{$\pm$ 0.0} \\

XLM-R\textbf{*} & 65.73 \tiny{$\pm$ 2.33} & 0.17 \tiny{$\pm$ 0.04} & 74.17 \tiny{$\pm$ 1.75} & 0.22 \tiny{$\pm$ 0.03} & 61.13 \tiny{$\pm$ 2.9} & 0.13 \tiny{$\pm$ 0.05} \\
RemBERT\textbf{*} & 76.21 \tiny{$\pm$ 0.33} & 0.44 \tiny{$\pm$ 0.01} & 78.32 \tiny{$\pm$ 0.75} & 0.40 \tiny{$\pm$ 0.02} & 75.06 \tiny{$\pm$ 0.55} & 0.44 \tiny{$\pm$ 0.01} \\ 

\midrule \multicolumn{7}{c}{\textbf{Decoder LMs (PenLP)}} \\ \midrule
ruGPT-3-small & 53.89 \tiny{$\pm$ 0.0} & 0.25 \tiny{$\pm$ 0.0} & 57.46 \tiny{$\pm$ 0.0} & 0.19 \tiny{$\pm$ 0.0} & 51.94 \tiny{$\pm$ 0.0} & 0.27 \tiny{$\pm$ 0.0}\\

ruGPT-3-medium & 55.79 \tiny{$\pm$ 0.0} & 0.27 \tiny{$\pm$ 0.0} & 59.39 \tiny{$\pm$ 0.0} & 0.19 \tiny{$\pm$ 0.0} & 53.82 \tiny{$\pm$ 0.0} & 0.30 \tiny{$\pm$ 0.0}\\

ruGPT-3-large & 56.83 \tiny{$\pm$ 0.0} & 0.29 \tiny{$\pm$ 0.0} & 61.22 \tiny{$\pm$ 0.0} & 0.22 \tiny{$\pm$ 0.0} & 54.43 \tiny{$\pm$ 0.0} & 0.31 \tiny{$\pm$ 0.0} \\

mGPT-XL\textbf{*} & 60.60 \tiny{$\pm$ 0.0} & 0.27 \tiny{$\pm$ 0.0} & 62.84 \tiny{$\pm$ 0.0} & 0.16 \tiny{$\pm$ 0.0} & 59.37 \tiny{$\pm$ 0.0} & 0.29 \tiny{$\pm$ 0.0} \\

\midrule

\multicolumn{7}{c}{\textbf{Encoder-decoder LMs}} \\ \midrule

ruT5-base & 71.26 \tiny{$\pm$ 1.31} & 0.27 \tiny{$\pm$ 0.03} & 76.49 \tiny{$\pm$ 1.54} & 0.33 \tiny{$\pm$ 0.03} & 68.41 \tiny{$\pm$ 1.55} & 0.25 \tiny{$\pm$ 0.04} \\

ruT5-large & 74.29 \tiny{$\pm$ 3.80} & 0.37 \tiny{$\pm$ 0.07} & 74.82 \tiny{$\pm$ 1.67} & 0.33 \tiny{$\pm$ 0.29} & 74.00 \tiny{$\pm$ 5.33} & 0.40 \tiny{$\pm$ 0.10} \\

FRED-T5-large & 75.83 \tiny{$\pm$ 0.0} & 0.40 \tiny{$\pm$ 0.0} & 77.36 \tiny{$\pm$ 0.0} & 0.34 \tiny{$\pm$ 0.0} & 75.0 \tiny{$\pm$ 0.0} & 0.42 \tiny{$\pm$ 0.0}  \\

FRED-T5-XL & 77.37 \tiny{$\pm$ 0.0} & 0.46 \tiny{$\pm$ 0.0} & 80.5 \tiny{$\pm$ 0.0} & 0.46 \tiny{$\pm$ 0.0} & 75.66 \tiny{$\pm$ 0.0} & 0.45 \tiny{$\pm$ 0.0} \\ \midrule


Human & \textbf{84.08}  & \textbf{0.63} &\textbf{83.55} &\textbf{0.57} &\textbf{84.59} & \textbf{0.67} \\ \bottomrule
\end{tabular}
\caption{Results for acceptability classification on the RuCoLA test set. The best score is in bold, and the second-best one is underlined. The baseline models are marked with an asterisk.}
\label{tab:rucola-results}
\vspace{-2ex}
\end{table*}

\paragraph{Results} The results are shown in~\autoref{tab:rsg_leaderboard}. FRED-T5-XL performs best on most tasks, with an overall score of $75.2$. Finetuning only the FRED-T5-XL encoder leads to strong results on PARus, MuSeRC, TERRa, RUSSE, and RuCoS. ruRoBERTa-large receives the overall best performance among the proposed encoder LMs ($68.1$), performing on par with ruT5-large. Comparing results with the best-performing encoder, we find that ruRoBERTa-large outperforms RuLeanALBERT on RCB and DaNetQA. We also find that our ruBERT-based LMs outperform DeepPavlov's ruBERT models. ruELECTRA performs worse on the machine reading comprehension tasks, which results in a lower overall score. The overall zero-shot performance of the decoder-only LMs is similar to the ruBERT-base-conv and ruELECTRA-based LMs. The larger versions of the ruGPT-based LMs outperform the encoders on RCB, PARus, and MuSeRC (e.g., mBERT, XLM-R-large, and ruELECTRA). 

Our LMs have promoted new state-of-the-art results on most of the Russian SuperGLUE tasks, and the overall performance gap between humans and the LMs has been narrowed by up to 4.9. However, there is still room for model improvement on the RWSD, RCB, TERRa, and PARus tasks.

\subsubsection{Acceptability Classification}
\label{sec:rucola}

\paragraph{Task} RuCoLA~\citeplanguageresource{mikhailov-etal-2022-rucola} consists of in-domain sentences from linguistic publications and out-of-domain sentences produced by generative LMs. The task is to predict if a given sentence is acceptable or not. The \textbf{performance metrics} are the accuracy score (Acc.) and MCC.

\paragraph{Method} We follow the finetuning and evaluation procedure described in~\citetlanguageresource{mikhailov-etal-2022-rucola}. We use the ruRoBERTa-large, ruGPT-3-medium, and ruT5-base results from~\citetlanguageresource{mikhailov-etal-2022-rucola}. The best model configuration is selected based on the MCC on the validation set. 

\begin{itemize}[leftmargin=1.1em,topsep=0.1em,itemsep=0em]
    \item Encoder LMs: the encoders (ruBERT, ruELECTRA) are finetuned for 5 epochs using the AdamW optimizer via a grid search over a set of hyperparameters: the learning rates $\{10^{-5}, 3\cdot 10^{-5}, 5\cdot 10^{-5}\}$ and the weight decay values $\{10^{-4}, 10^{-2}, 0.1\}$. The results are averaged over 10 experiment runs with different random seeds. 
    
    \item Decoder LMs: the ruGPT-3-small and ruGPT-3-large models are evaluated using a classification approach based on a threshold for the PenLP acceptability measure~\cite{lau-etal-2020-furiously}. The threshold is selected on the training set via 10-fold cross-validation to maximize MCC on the validation set: $-19.65$ (ruGPT-3-small), $-20.91$ (ruGPT-3-medium), and $-19.39$ (ruGPT-3-large).

    \item Encoder-decoder LMs: ruT5-large is finetuned for 20 epochs, with the search space of $\{10^{-4}, 10^{-3}\}$ for the learning rate and $\{0, 10^{-4}\}$ for the weight decay. We finetune the FRED-T5 models for 20 epochs using the Adafactor optimizer,  the learning rate of $5 \cdot 10^{-4}$, weight decay of $0.0$, and batch size of $16$. We report the results for only one experiment run.
\end{itemize}

 \paragraph{Baselines} We finetune ruBERT-base by DeepPavlov, ruBERT-base-conv, and mBERT as described above. The PenLP threshold for mGPT-XL\footnote{\href{https://huggingface.co/ai-forever/mGPT}{\texttt{hf.co/ai-forever/mGPT}}} is $-54.37$. We use the results for human annotators, XLM-R, and RemBERT from~\citetlanguageresource{mikhailov-etal-2022-rucola}. Results for RuLeanALBERT are from the RuCoLA leaderboard as of September 2023: \href{https://rucola-benchmark.com/leaderboard}{\texttt{rucola-benchmark.com/leaderboard}}.

 \paragraph{Results} The results for acceptability classification are presented in~\autoref{tab:rucola-results}. In general, our LMs outperform their monolingual and multilingual counterparts. ruRoBERTa-large receives the best performance among the LMs, falling short behind expert human annotators. The second-best is RuLeanALBERT, followed by FRED-T5-XL and RemBERT. At the same time, ruELECTRA outperforms mBERT and XLMR. We observe that ruGPT-3-large performs the best among the threshold-based classifiers, and the ruGPT-3-medium performance is similar to mGPT 1.3B. Our LMs generalize well to machine-generated sentences, showing minor performance differences between the in- and out-of-domain sets.

\begin{table}[t!]
 \centering
    \scriptsize
\begin{tabular}{@{}lc@{}}
\toprule
\textbf{Model} & \textbf{F1-score} \\
\midrule 
\multicolumn{2}{c}{\textbf{Encoder LMs}} \\
\midrule
ruBERT-base  & 80.75 \tiny{$\pm$ 0.32} \\
ruBERT-large & 81.27 \tiny{$\pm$ 0.34}\\
ruRoBERTa-large & \underline{82.44} \tiny{$\pm$ 1.02} \\
ruELECTRA-small & 78.46 \tiny{$\pm$ 0.77} \\ 
ruELECTRA-medium & 79.05 \tiny{$\pm$ 0.43}  \\
ruELECTRA-large & 80.27 \tiny{$\pm$ 1.30} \\
mBERT\textbf{*} & 78.24 \tiny{$\pm$ 0.56} \\ 
ruBERT-base (DP)\textbf{*} & 79.59 \tiny{$\pm$ 0.07} \\ 
ruBERT-base-conv (DP)\textbf{*} & 81.14 \tiny{$\pm$ 0.64} \\ 
\midrule 
\multicolumn{2}{c}{\textbf{Decoder LMs}} \\
\midrule
ruGPT-3-small & 64.68 \tiny{$\pm$ 0.0} \\
ruGPT-3-medium & 64.32 \tiny{$\pm$ 0.0} \\ 
ruGPT-3-large & 64.39 \tiny{$\pm$ 0.0} \\ 
mGPT-XL\textbf{*} & 64.78 \tiny{$\pm$ 0.0} \\ 
\midrule
\multicolumn{2}{c}{\textbf{Encoder-decoder LMs}}\\
\midrule
ruT5-base & 75.45 \tiny{$\pm$ 0.0} \\
ruT5-large & 75.20 \tiny{$\pm$ 0.0}\\
FRED-T5-large & 82.13 \tiny{$\pm$ 0.0} \\
FRED-T5-XL & \textbf{82.86} \tiny{$\pm$ 0.0}\\
mT5-base\textbf{*} & 75.63 \tiny{$\pm$ 0.0} \\ 
mT5-large\textbf{*} & 77.33 \tiny{$\pm$ 0.0} \\ 
\bottomrule
\end{tabular}
\caption{Results for inappropriateness identification. DP=DeepPavlov~\cite{burtsev-etal-2018-deeppavlov}. The best score is in bold, and the second best is underlined. The baseline models are marked with an asterisk.
}
\label{tab:ii}
\end{table}

\subsubsection{Inappropriateness Identification}
\label{sec:ii}
\paragraph{Task} We use the dataset by~\citetlanguageresource{babakov-etal-2021-detecting} to evaluate the model's ability to identify inappropriate messages, which can cover a sensitive topic (e.g., crime, body shaming, and sexism) and harm the reputation of the user. The target \textbf{performance metric} is the macro-average F1-score. 

 \paragraph{Method} We finetune and evaluate the encoder, decoder, and encoder-decoder LMs as described in \S\ref{sec:rucola}. The PenLP thresholds are $-37.66$ (ruGPT-3-small), $-35.82$ (ruGPT-3-medium), and $-35.39$ (ruGPT-3-large).

 \paragraph{Baselines} We finetune and evaluate mBERT, ruBERT-base by DeepPavlov, ruBERT-base-conv, mT5-base, and mT5-large as described in~\S\ref{sec:rucola}. The PenLP threshold for mGPT-XL is $-32.54$.

 \paragraph{Results}
The results for inappropriateness identification are presented in~\autoref{tab:ii}. Overall, all models receive strong performance, and the encoder and decoder-only LMs perform on par. The performance improves with the model scaling, except for the decoder-only and ruT5 models. FRED-T5-XL shows the best results among the LMs, followed by ruRoBERTa-large and FRED-T5-large. 

\subsection{Natural Language Generation}

\subsubsection{Text Simplification}
\label{sec:simplification}
\paragraph{Task} RuSimpleSentEval-2021~\citelanguageresource{sakhovskiy2021rusimplesenteval} is a corpus of pairs of sentences comprising complex sentences and their simplified versions. The task is to rewrite the input sentence in a less complicated way. The \textbf{performance metrics} are 
SARI~\cite{xu-etal-2015-problems} and BERTScore~\cite{zhang2019bertscore} computed between the input and the output using mBERT. 

\paragraph{Method}  We finetune the decoder and encoder-decoder LMs using the AdamW optimizer, the learning rate of $5 \cdot 10^{-5}$, and batch size of $2$ for $3$ and $10$ epochs, respectively. The decoding strategy and hyperparameters for inference are selected based on the validation performance and manual analysis of the model outputs. The resulting strategy is beam search with $5$ beams for all models.

 \paragraph{Baselines} We report human reference scores
and a non-neural baseline of the input sentence without any change (Input sentence). Then, following the procedure described above, we finetune mBART-large-50~\cite{tang-etal-2021-multilingual}, mGPT-XL, mT5-base, and mT5-large.

\begin{table}[t!]
    \centering
    \tiny 
    \resizebox{\columnwidth}{!}{
\begin{tabular}{@{}lcccc@{}}
\toprule
\multirowcell{2}[-0.5ex][l]{\textbf{Model}} & \multicolumn{2}{c}{\textbf{Public test}} & \multicolumn{2}{c}{\textbf{Private test}}  \\ \cmidrule(lr){2-3} \cmidrule(lr){4-5} 
 & \textbf{SARI} & \textbf{BERTScore} & \textbf{SARI} & \textbf{BERTScore}  \\ 





\midrule
\multicolumn{5}{c}{\textbf{Decoder LMs}} \\ 

\midrule

ruGPT-3-small & 37.96  & 0.81  & 37.54  & 0.79  \\
ruGPT-3-medium & 39.00  & 0.91  & 39.21  & 0.91 \\
ruGPT-3-large & 39.09  & 0.90  & 39.37  & 0.90 \\
mGPT-XL\textbf{*}& 42.45 & 0.98  & 42.22   & 0.97   \\


\midrule \multicolumn{5}{c}{\textbf{Encoder-decoder LMs}} \\ \midrule

ruT5-base & 43.34  & \underline{1.0}  & 43.29  & \underline{1.0} \\
ruT5-large & 43.33 & \underline{1.0}  & 43.22  & \underline{1.0} \\
FRED-T5-large & \underline{43.95}  & 0.99  & 43.40  & 0.99 \\
FRED-T5-XL &  43.41  & \underline{1.0}  & 43.35  & 0.99 \\
mBART-large-50\textbf{*} & 39.75 & 0.95  & 40.47   & 0.96   \\
mT5-base\textbf{*} & 43.63 & 0.99  & 43.55   & 0.99   \\
mT5-large\textbf{*}  & 43.62 & \underline{1.0}  & 43.68   & \underline{1.0}   \\







\midrule

Input sentence\textbf{*}  & 43.90 & \underline{1.0}  & \underline{43.92}  & \underline{1.0}   \\ \midrule

Human  & \textbf{66.72} & 0.82   & \textbf{66.11 }  & 0.82   \\


\bottomrule
\end{tabular}}
\caption{Results for text simplification on the RuSimpleSentEval-2021 test sets. The best score is in bold, and the second best one is underlined. The baseline models are marked with an asterisk.}
\label{tab:simplification}
\end{table}

 \paragraph{Results}
The results for the text simplification task are presented in~\autoref{tab:simplification}. For all tested models except for ruGPT3-small, BERTScore exceeds 0.9, which means that simplified predictions are very close to the input sentence with slight simplifications, mainly at the word level. Overall, our manual analysis of the model outputs suggests that the target metric (SARI) does not indicate the intended performance. For instance, the multilingual LMs (mT5 and mBART-large-50) tend to copy most parts of the input, which results in high BERTScore (over $0.96$) and strong SARI scores. At the same time, SARI does not always improve with the model scaling. We also find that encoder-decoder LMs outperform decoder-only LMs, and ruT5-base leaves the input sentence unchanged, similar to mT5 and mBART-large-50. The results indicate that it is necessary to conduct a human-based evaluation to get a more complete picture of the model performance.

\subsubsection{Text Summarization}
\label{sec:summarization}
\paragraph{Task} Gazeta~\citelanguageresource{gusev2020dataset} is a corpus of news articles and their summaries for abstractive summarization. The \textbf{performance metrics} are standard summarization evaluation metrics: ROUGE-L~\cite{lin-2004-rouge}, BERTScore, BLEU~\cite{papineni-etal-2002-bleu}, METEOR~\cite{banerjee-lavie-2005-meteor}, and ChrF1~\cite{popovic-2015-chrf}. 

 \paragraph{Method} 
We finetune the decoder-only models for 3 epochs using AdamW optimizer, a linear scheduler with a warmup, and a learning rate of $5 \cdot 10^{-5}$. The encoder-decoder models are finetuned with Adafactor with a constant learning rate of $1 \cdot 10^{-3}$. We examine different generation strategies and hyperparameters on the validation set. The resulting strategy is beam search with $5$ beams for all LMs.

\paragraph{Baselines} 
We finetune mBART-large-50, mT5-base, and mT5-large as described above. 

\begin{table}[t!]
    \centering
    \tiny 
    \resizebox{\columnwidth}{!}{
\begin{tabular}{@{}lccccc@{}}
\toprule
\textbf{Model} & \textbf{ROUGE-L}  & \textbf{BERTScore}  & \textbf{BLEU}  & \textbf{METEOR} &  \textbf{ChrF1} \\ 

\midrule 
\multicolumn{6}{c}{\textbf{Decoder LMs}} \\ 
\midrule
ruGPT-3-small & 17.28 & 71.78 & 6.18 & 20.13 &  30.66\\
ruGPT-3-medium & 19.27 & 72.37 & 6.89 & 21.81 & 32.72\\
ruGPT-3-large & 19.66 & 72.62 & 7.24 & 22.39 &33.37\\

\midrule 
\multicolumn{6}{c}{\textbf{Encoder-decoder LMs}} \\ 
\midrule

ruT5-base & 18.72 & 73.15 & 7.42 & 22.78 &  33.17\\
ruT5-large & 20.12 & 73.53 & 8.11 & 23.9 & 34.59\\
FRED-T5-large & \underline{22.48} & \underline{73.69} & \underline{8.35} & \underline{24.29}  & \underline{34.97}\\
FRED-T5-XL & \textbf{22.95} & \textbf{73.9} & \textbf{8.61} & \textbf{24.72} & \textbf{35.36} \\
mBART-large-50\textbf{*}  & 18.53  & 72.58  & 7.46 & 22.63 & 34.95 \\
mT5-base\textbf{*}  & 17.76  & 71.96  & 6.16 & 20.45 & 30.95 \\
mT5-large\textbf{*}  & 17.80  & 72.73  & 7.16 & 21.84 & 33.38 \\
\bottomrule
\end{tabular}}
\caption{Results for text summarization on Gazeta. The best score is in bold, second best is underlined. The baseline models are marked with an asterisk.}
\label{tab:summarization}
\vspace{-2ex}
\end{table}

\paragraph{Results} The results for text summarization are shown in~\autoref{tab:summarization}. The scores demonstrate that the performance improves as the model size increases. ruGPT-3-large achieves the highest scores among the decoder LMs, and FRED-T5-XL receives the best performance among the encoder-decoder LMs. The manual analysis of the model outputs indicates that the ruGPT-3 models tend to copy parts of the inputs, while the ruT5 and FRED-T5 models produce more plausible summaries. Overall, our LMs show higher scores as opposed to their multilingual counterparts.

\subsubsection{Text Detoxification}
\label{sec:detoxification}
\paragraph{Task} The RUSSE Detoxification corpus~\citelanguageresource{dementievarusse} tests the model's capability of generating a detoxified version of the toxic text. The \textbf{performance metrics} are based on~\citetlanguageresource{dementievarusse}: ChrF1 score, style transfer accuracy, content similarity,  fluency, and the ``Joint'' score (multiplication of last three metrics).

\paragraph{Method} We conduct finetuning of the LMs over five epochs using AdamW for the rGPT-based models and Adafactor for the ruT5-based models. We experiment with multiple decoding strategies on the validation set, analyzing the performance metrics and conducting manual analysis of the outputs. We use beam search with $5$ beams and the repetition penalty of $1.05$ at the inference stage.

 \paragraph{Baselines.}
We report human reference scores and baseline results provided by~\citetlanguageresource{dementievarusse}: (i) a trivial ``Duplicate'' baseline, which leaves the original text intact and acts as a lower performance threshold; (ii) a ``Delete'' baseline, which removes toxic words based on a predefined vocabulary. Additionally, we finetune and evaluate mBART-large-50, mT5-base, and mT5-large with the same parameters as the LMs above.

\begin{table}[t!]
    \centering
    \tiny 
    \resizebox{\columnwidth}{!}{\begin{tabular}{@{}lccccc@{}}
\toprule
 \textbf{Model} & \textbf{STA}  & \textbf{SIM}  & \textbf{FL}  & \textbf{Joint} & \textbf{ChrF1} \\ \midrule
 \multicolumn{6}{c}{\textbf{Decoder LMs}} \\ \midrule
ruGPT-3-small & 74.0 & 80.2 & 83.5 & 50.4 & 51.8 \\
ruGPT-3-medium & 78.0 & 79.8 & 83.6 & 53.1 & 54.0 \\
ruGPT-3-large & 75.4 & 81.4 & 82.6 & 50.8 & 55.5\\

\midrule
\multicolumn{6}{c}{\textbf{Encoder-decoder LMs}} \\
\midrule
ruT5-base & 80.0 & 81.9 & 83.0 & 55.3 & 57.2\\
ruT5-large & 78.8 & 81.6 & 83.2 & 54.4 & 56.8\\
FRED-T5-large & 81.9 & 81.8 & 84.8 & \underline{57.8} & 57.6\\
FRED-T5-XL & \underline{82.3} & 82.1 & 85.3 & \textbf{58.5} & \underline{58.1}\\
mBART-large-50\textbf{*} & 81.4  & 77.5  & 79.7  & 51.5 & 53.6 \\
mT5-base\textbf{*} & 61.5 &  86.4 & 83.1  & 42.8 & 54.9 \\
mT5-large\textbf{*} & 77.4  & 84.5  & \underline{86.1}  & 56.7 & 56.9 \\

\midrule
Duplicate\textbf{*} & 24.0 & \textbf{100.0} & \textbf{100.0} & 24.0 & 56.0 \\ \midrule
Delete\textbf{*} & 55.8  & \underline{88.7}  & 85.2  & 40.6 & 52.6 \\

\midrule
Human & \textbf{85.0} & 72.0 & 78.0 & 49.0 & \textbf{77.0} \\

\bottomrule
\end{tabular}}
\caption{Results for detoxification. Performance metrics: STA=Style Transfer Accuracy, SIM=Content Similarity, FL=Fluency. The best score is in bold, second best is underlined. The baseline models are marked with an asterisk.}
\label{tab:detox}
\end{table}

 \paragraph{Results}
The text detoxification results are presented in~\autoref{tab:detox}. The scores show that the LMs demonstrate a significant performance improvement over the baselines when considering the ``Joint'' score and surpass human performance with regard to text similarity and fluency. The performance difference between the decoder-only and encoder-decoder LMs is not substantial. However, the encoder-decoder LMs perform better, with FRED-T5-XL achieving the highest Joint score ($58.5$) and the best model ChrF1 score ($58.1$).

\section{Conclusion}
This paper introduces 13 Russian Transformer LMs of various model architectures, pretraining objectives, and model sizes. We have released our LMs over the last few years, facilitating research advancements and the development of specialized downstream solutions for the Russian language. We provide a report on the model architecture design, pretraining corpus, and pretraining. We empirically evaluate our LMs, their multilingual counterparts, and other open-source Russian LMs on standard Russian NLP benchmarks and datasets. The results indicate that our LMs promote state-of-the-art performance on Russian SuperGLUE and RuCoLA and match the human performance on the machine reading comprehension and text detoxification tasks. We outline the following future work research directions that are out of the scope of this paper: (i)
analyzing the model performance when finetuning data is limited, (ii) exploring the effect of pretraining corpus composition, (iii) other techniques for adapting language models to Russian, such as initializing from a multilingual LM, (iv) conducting a more optimal hyperparameter search, and (v) performing a human-based generation evaluation. We aim to continue to develop novel Russian LMs in the future.

\section{Limitations}
\paragraph{Limited Context Size} Although our generative LMs achieve strong results and promote state-of-the-art performance on various tasks, their context window size (maximum 2048 tokens) limits the model application on long-context tasks. We leave experiments with efficient finetuning approaches to extending the context size for future work~(\citealp[e.g.,][]{chen2023longlora}).

\paragraph{Social Bias Evaluation} The evaluation experiments conducted in this paper do not -- and de facto cannot -- address all possible scenarios. We aim to assess our model generalization abilities on standard academic datasets and benchmarks, covering various natural language understanding and generation tasks. Still, our experimental setup is limited due to the lack of peer-reviewed resources for specific evaluation cases, such as detecting social biases, stereotypes, and hate speech. Therefore, before deploying our LMs, developers should perform safety evaluations for their specific model application scenarios. 

\paragraph{Language Generation Evaluation} The performance metrics for natural language generation tasks do not always capture the task-specific properties~(\citealp[e.g.,][]{fomicheva-specia-2019-taking,colombo2022glass,chhun-etal-2022-human}). Our manual analysis of the model outputs confirms these findings for the text simplification task (see~\S\ref{sec:simplification}). While we follow the evaluation approach based on a combination of standard performance metrics of different types, these metrics may not comprehensively evaluate the model generation abilities. We suggest a human-based side-by-side model evaluation may help get a complete picture of the performance.

\paragraph{Domain Shifts} Our LMs' pretraining corpus features various domains, including general domain, news, books, web texts, and subtitles. However, pretraining the LMs\footnote{Recall that our LMs have been pretrained over the last several years, and the domain choice and sub-corpora sizes are based on multiple factors (see \S\ref{sec:pretrain}).} on different sub-corpora can hinder their performance in domain-specific applications and on out-of-domain data. Nevertheless, we empirically show that our LMs receive strong performance on domains not well represented in the pretraining corpus, ranging from linguistic publications (\S\ref{sec:rucola}) to user messages (\S\ref{sec:ii}).

\section{Ethical Considerations}
\label{sec:ethicalstatement}
The development of the new LMs detailed in this paper adheres to standard ethical guidelines. We advocate for these models' responsible and impartial utilization, carefully considering their potential societal impacts. Special attention is given to filtering harmful content and ensuring a diverse range of perspectives and sources are included in the model pretraining corpora. Furthermore, we recognize the importance of ongoing vigilance in monitoring and addressing the unintended consequences of deploying these models in real-world applications.

 \paragraph{Possible Misuse} We believe that our research should not be involved in creating content that somehow affects the individual or communal well-being, including (i) legislative application or censorship, (ii) disinformation, infringement of the rights of access to information, (iii) dehumanizing, misrepresenting, or otherwise harmful representations of people or their religions, culture, belief, (iv) promoting harmful or discriminatory content.

 \paragraph{Biases and data quality} The pretraining data for some of the presented models includes large segments from the internet domain and, consequently, contains various stereotypes and biases. Therefore, proper model evaluation is still needed to explore their possible vulnerabilities in generalizing to the out-of-domain data.

 \paragraph{Energy Efficiency and Usage} We compute the $CO_2$ emissions from pretraining our LMs as Equation~\ref{eq:co2}~\cite{strubelletal2019energy}:

\vspace{-2pt}
\begin{equation}
\label{eq:co2}
    CO_2 = \frac{PUE * kWh * I^{CO2}}{1000}
\end{equation}

\begin{table}[t!]
    \small
    \centering
    \setlength{\tabcolsep}{2pt}
    \begin{tabular}{lc}
    \toprule
       \textbf{Model} & \textbf{$CO_2$ (kg)} \\
       
       \midrule
       \multicolumn{2}{c}{\textbf{Encoder LMs}} \\
       \midrule
        ruBERT-base & 1.17k \\
        ruBERT-large & 2.94k \\
        ruRoBERTa-large & 12.37k \\
        ruELECTRA-small & 0.25k \\ 
        ruELECTRA-medium & 0.29k \\
        ruELECTRA-large & 0.36k \\

       \midrule
       \multicolumn{2}{c}{\textbf{Encoder-decoder LMs}} \\
       \midrule
       ruT5-base & 4.12k \\
       ruT5-large & 12.37k \\

       FRED-T5-large & 55.7k \\
       FRED-T5-XL & 52.7k \\

       \midrule
       \multicolumn{2}{c}{\textbf{Decoder LMs}} \\
       \midrule

       ruGPT-3-small & 2.06k \\
       ruGPT-3-medium & 9.43k \\
       ruGPT-3-large & 16.94k \\

    \bottomrule
    
    \end{tabular}
    \caption{$CO_2$ emissions of pretraining models.}
    \label{tab:co2}
\end{table}

\noindent The power usage effectiveness ($PUE$) of our data centers is $1.3$. The $CO_2$ emissions in kg are presented in~\autoref{tab:co2}. Model compression techniques and parameter-efficient finetuning methods can reduce the computational costs associated with model inference. Note that while the ruELECTRA models underperform the baselines on some natural language understanding tasks (e.g., machine reading comprehension), these LMs are highly efficient due to their size (e.g., the small and medium versions have 42M and 85M, respectively). We recommend the user conduct their own evaluation for a downstream task of interest accounting for both performance and efficiency. 


\nocite{*}
\section{Bibliographical References}\label{sec:reference}
\bibliographystyle{lrec-coling2024-natbib}
\bibliography{lrec-coling2024-example}

\section{Language Resource References}
\label{lr:ref}
\bibliographystylelanguageresource{lrec-coling2024-natbib}
\bibliographylanguageresource{languageresource}

\clearpage
\newpage

\onecolumn

\section{Appendix}
\label{sec:appendix}

\subsection{Hyperparameter Values}
\label{app:hparams}

\begin{table}[h!]
\centering
    \small
    
\begin{tabular}{lcccc}
\toprule
 \textbf{Model} & \textbf{Optimizer} & \textbf{Learning Rate} &  \textbf{Weight Decay}  & \textbf{Batch Size} \\

\midrule
\multicolumn{5}{c}{\textbf{Russian SuperGLUE}} \\
\midrule
Encoder LMs & AdamW & $ 1\cdot10^{-5}$ & $0.01$ & $32$ \\
Decoder LMs & \xmark & \xmark  & \xmark & \xmark \\
Encoder-decoder LMs (I) &  Adafactor & $1\cdot10^{-3}$ & $15$ & $16$\\
Encoder-decoder LMs (II) & Adam & $1\cdot10^{-5}$ & $20$ & $16$\\

\midrule
\multicolumn{5}{c}{\textbf{RuCoLA}}\\
\midrule
ruBERT-base & AdamW & $3 \cdot 10^{-5}$ & $1e^{-4}$ & $64$ \\
ruBERT-large & AdamW & $3 \cdot 10^{-5}$ & $0.1$ & $64$ \\
ruBERT-base (DP) & AdamW  & $3 \cdot 10^{-5}$ & $0.01$ & $64$\\
ruBERT-base-conv (DP)  & AdamW & $1 \cdot 10^{-5}$ & $0.01$ & $32$ \\ 
mBERT  & AdamW & $1 \cdot 10^{-5}$ & $0.1$ & $32$  \\ 
ruRoBERTa-large & AdamW & $10^{-5}$ & $10^{-4}$ & $32$ \\
ruELECTRA-small & AdamW & $5 \cdot 10^{-5}$ & $0.1$ & $32$ \\
ruELECTRA-medium & AdamW & $5 \cdot 10^{-5}$ & $0.1$ & $32$ \\
ruELECTRA-large & AdamW & $3 \cdot 10^{-5}$ & $10^{-4}$ & $32$ \\
ruT5-base & Adafactor & $10^{-4}$ & $0$&  $128$ \\
ruT5-large & Adafactor & $10^{-4}$ & $0$&  $128$ \\
FRED-T5-large & Adafactor & $5 \cdot 10^{-4}$ & $0$&  $16$ \\
FRED-T5-XL & Adafactor & $5 \cdot 10^{-4}$ & $0$&  $16$
\\ 

\midrule
\multicolumn{5}{c}{\textbf{Inappropriateness Identification}}\\
\midrule
ruBERT-base & AdamW & $1 \cdot 10^{-5}$ & $0.1$ & $64$ \\
ruBERT-large & AdamW & $1 \cdot 10^{-5}$ & $0.1$ & $16$ \\
ruBERT-base (DP) & AdamW  & $1 \cdot 10^{-5}$ & $0.1$ & $64$\\
ruBERT-base-conv (DP)  & AdamW & $1 \cdot 10^{-5}$ & $0.01$ & $64$ \\ 
mBERT  & AdamW & $3 \cdot 10^{-5}$ & $0.01$ & $32$  \\ 
ruRoBERTa-large & AdamW & $10^{-5}$ & $10^{-4}$ & $32$ \\
ruELECTRA-small & AdamW & $5 \cdot 10^{-5}$ & $10^{-3}$ & $64$ \\
ruELECTRA-medium & AdamW & $5 \cdot 10^{-5}$ & $0.01$ & $64$ \\
ruELECTRA-large & AdamW &  &  &  \\
ruT5-base & Adafactor & $10^{-4}$ & $0$&  $128$ \\
ruT5-large & Adafactor & $10^{-4}$ & $0$&  $128$ \\
FRED-T5-large & Adafactor & $5 \cdot 10^{-4}$ & $0$&  $16$ \\
FRED-T5-XL & Adafactor & $5 \cdot 10^{-4}$ & $0$&  $16$

 \\ \midrule
\multicolumn{5}{c}{\textbf{Text Simplification}}\\
\midrule
Decoder LMs & AdamW & $1\cdot10^{-5}$ & $0$ & $2$ \\
Encoder-decoder LMs & AdamW & $1\cdot10^{-5}$ & $0$ & $2$ \\\midrule
\multicolumn{5}{c}{\textbf{Text Detoxification}}\\
\midrule
Decoder LMs & AdamW & $5 \cdot 10^{-5}$ & $0.01$ & $2$ \\
Encoder-decoder LMs & Adafactor & $1\cdot10^{-4}$ & $0.01$ & $8$ \\

\midrule
\multicolumn{5}{c}{\textbf{Text Summarization}}\\
\midrule
Decoder LMs & AdamW & $5 \cdot 10^{-5}$ & $0.01$ & $4$ \\
Encoder-decoder LMs & Adafactor & $1\cdot10^{-3}$ & $0.01$ & $2$ \\

\bottomrule
\end{tabular}
\caption{Optimal hyperparameter values found in the experiments. I/II=finetuning stage. DP=DeepPavlov~\cite{burtsev-etal-2018-deeppavlov}.}
\label{tab:hparams_finetune}
\end{table}

\clearpage
\newpage
\onecolumn

\begin{table*}[h!]
    \subsection{Russian SuperGLUE Templates}
    \label{app:rsg_templates}
    \centering
    \resizebox{\textwidth}{!}{
    \begin{tabular}{llc}
        \toprule
        \textbf{Model} & \textbf{Format} & \textbf{Labels} \\

        \midrule
        \multicolumn{3}{c}{\textbf{LiDiRus}} \\
        \midrule

        ruRoBERTa &\texttt{<s> \lb premise\rb\ </s></s> \lb hypothesis\rb\ </s>} &  \small \texttt{entailment | not\_entailment }\\
        ruBERT &\texttt{[CLS] \lb premise\rb\ [SEP] \lb hypothesis\rb\ [SEP]} & \small \texttt{entailment | not\_entailment}  \\
        ruELECTRA & \texttt{[CLS] \lb premise\rb\ [SEP] \lb hypothesis\rb\ [SEP]} & \small \texttt{entailment | not\_entailment} \\
        ruT5 
            &  \texttt{lidirus premise: \lb premise\rb\ hypothesis: \lb hypothesis\rb\ } & \small \texttt{entails | doesn't entail} \\
        FRED-T5 
            &  \texttt{lidirus premise: \lb premise\rb\ hypothesis: \lb hypothesis\rb\ } & \small \texttt{entails | doesn't entail} \\
        
        \midrule
        \multicolumn{3}{c}{\textbf{RCB}} \\
        \midrule

        ruRoBERTa &\texttt{<s> \lb premise\rb\ </s></s> \lb hypothesis\rb\ </s>} &  \small \texttt{entailment | contradiction | neutral}\\
        ruBERT &\texttt{[CLS] \lb premise\rb\ [SEP] \lb hypothesis\rb\ [SEP]} &  \small \texttt{entailment | contradiction | neutral}\\
        ruELECTRA &\texttt{[CLS] \lb premise\rb\ [SEP] \lb hypothesis\rb\ [SEP]} &  \small \texttt{entailment | contradiction | neutral}\\
        ruT5 
            &  \texttt{rcb premise: \lb premise\rb\ hypothesis: \lb hypothesis\rb\ } & \small \texttt{entailment | contradiction | neutral}\\
        FRED-T5 
            &  \texttt{rcb premise: \lb premise\rb\ hypothesis: \lb hypothesis\rb\ } & \small \texttt{entailment | contradiction | neutral}\\

        \midrule
        \multicolumn{3}{c}{\textbf{PARus}} \\
        \midrule

        ruRoBERTa &\texttt{<s> \lb premise\rb\ </s></s> \lb hypothesis\rb\ </s>} &  \small \texttt{0 | 1}\\
        ruBERT &\texttt{[CLS] \lb premise\rb\ [SEP] \lb hypothesis\rb\ [SEP]} &  \small \texttt{0 | 1}\\
        ruELECTRA &\texttt{[CLS] \lb premise\rb\ [SEP] \lb hypothesis\rb\ [SEP]} &  \small \texttt{0 | 1}\\
        ruT5 
            &  \texttt{parus premise: \lb premise\rb\ hypothesis1: \lb choice1\rb\ hypothesis2: \lb choice2\rb\ } & \small \texttt{hypothesis1 | hypothesis2} \\
        FRED-T5 
            &  \texttt{parus premise: \lb premise\rb\ hypothesis1: \lb choice1\rb\ hypothesis2: \lb choice2\rb\ } & \small \texttt{hypothesis1 | hypothesis2} \\

        \midrule
        \multicolumn{3}{c}{\textbf{MuSeRC}} \\
        \midrule

        ruRoBERTa &\texttt{<s> \lb passage\rb\ </s></s> \lb question\rb\ \lb answer\rb\ </s>} &  \small \texttt{0 | 1}\\
        ruBERT &\texttt{[CLS] \lb passage\rb\ [SEP] \lb question\rb\ \lb answer\rb\ [SEP]} &  \small \texttt{0 | 1}\\
        ruELECTRA &\texttt{[CLS] \lb passage\rb\ [SEP]\lb question\rb\ \lb answer\rb\ [SEP]} &  \small \texttt{0 | 1}\\
        ruT5 
            &  \texttt{muserc question: \lb question\rb\ answer: \lb answer\rb\ text: \lb passage\rb\ } & \small \texttt{no | yes} \\
        FRED-T5 
            &  \texttt{muserc question: \lb question\rb\ answer: \lb answer\rb\ text: \lb passage\rb\ } & \small \texttt{no | yes} \\

        \midrule
        \multicolumn{3}{c}{\textbf{TERRa}} \\
        \midrule

        ruRoBERTa &\texttt{<s> \lb premise\rb\ </s></s> \lb hypothesis\rb\ </s>} & \small \texttt{entailment | not\_entailment} \\
        ruBERT &\texttt{[CLS] \lb premise\rb\ [SEP] \lb hypothesis\rb\ [SEP]} & \small \texttt{entailment | not\_entailment}  \\
        ruELECTRA &\texttt{[CLS] \lb premise\rb\ [SEP] \lb hypothesis\rb\ [SEP]} & \small \texttt{entailment | not\_entailment} \\
        ruT5 
            &  \texttt{terra premise: \lb premise\rb\ hypothesis: \lb hypothesis\rb\ } & \small \texttt{entails | doesn't entail} \\
        FRED-T5 
            &  \texttt{terra premise: \lb premise\rb\ hypothesis: \lb hypothesis\rb\ } & \small \texttt{entails | doesn't entail} \\

        \midrule
        \multicolumn{3}{c}{\textbf{RUSSE}} \\
        \midrule

        ruRoBERTa &\texttt{<s> \lb sentence1\rb\ </s></s> \lb sentence2\rb\ </s></s> \lb word\rb\ </s>} &  \small \texttt{True | False}\\
        ruBERT &\texttt{[CLS] \lb sentence1\rb\ [SEP] \lb sentence2\rb\ [SEP]} &   \small \texttt{True | False}\\
        ruELECTRA &\texttt{[CLS] \lb sentence1\rb\ [SEP] \lb sentence2\rb\ [SEP]} &  \small \texttt{True | False} \\
        ruT5 
            &  \texttt{russe sentence1: \lb sentence1\rb\ sentence2: \lb sentence2\rb\ slovo: \lb word\rb} & \small \texttt{no | yes} \\
        FRED-T5 
            &  \texttt{russe sentence1: \lb sentence1\rb\ sentence2: \lb sentence2\rb\ slovo: \lb word\rb} & \small \texttt{no | yes} \\

        \midrule
        \multicolumn{3}{c}{\textbf{RWSD*}} \\
        \midrule

        ruRoBERTa &\texttt{False} &  \\
        ruBERT &\texttt{False} &  \\
        ruELECTRA &\texttt{False} &  \\
        ruT5 
            &  \texttt{False} &  \\
        FRED-T5 
            &  \texttt{False} &  \\

        \midrule
        \multicolumn{3}{c}{\textbf{DaNetQA}} \\
        \midrule

        ruRoBERTa &\texttt{<s> \lb passage\rb\ </s></s> \lb question\rb\ </s>} &  \small \texttt{0 | 1}\\
        ruBERT &\texttt{[CLS] \lb passage\rb\ [SEP] \lb question\rb\ [SEP]} &  \small \texttt{0 | 1}\\
        ruELECTRA &\texttt{[CLS] \lb passage\rb\ [SEP] \lb question\rb\ [SEP]} &  \small \texttt{0 | 1}\\
        ruT5 
            &  \texttt{danetqa question: \lb question\rb\ text: \lb passage\rb} & \small \texttt{no |  yes} \\
        FRED-T5 
            &  \texttt{danetqa question: \lb question\rb\ text: \lb passage\rb} & \small \texttt{no |  yes} \\

        \midrule
        \multicolumn{3}{c}{\textbf{RuCoS}} \\
        \midrule

        ruRoBERTa &\texttt{<s> \lb passage\rb\ </s></s> \lb query.replace('@placeholder', entities[i])\rb\ </s>} & \small \texttt{0 | 1} \\
        ruBERT &\texttt{[CLS] \lb passage\rb\ [SEP] \lb query.replace('@placeholder', entities[i])\rb\ [SEP]} &  \small \texttt{0 | 1}\\
        ruELECTRA &\texttt{[CLS] \lb passage\rb\ [SEP] \lb query.replace('@placeholder', entities[i])\rb\ [SEP]} & \small \texttt{0 | 1} \\
        ruT5 
            &  \texttt{rucos question: \lb query\rb\ entities: \lb ', '.join(entities)\rb} & \small  \texttt{\lb entities[i]\rb} \\
        FRED-T5 
            &  \texttt{danetqa question: \lb question\rb\ text: \lb passage\rb} & \small \texttt{\lb entities[i]\rb} \\

        \bottomrule
    \end{tabular}}
    \caption{Example templates for the RussianSuperGLUE tasks. * -- due to the task complexity, we submit the majority baseline for the RWSD task as our best performing model.}
    \label{tab:rsg_prompts}
\end{table*}
\end{document}